\documentclass{article}

     \usepackage[preprint]{neurips_2020}

\usepackage[utf8]{inputenc} 
\usepackage[T1]{fontenc}    
\usepackage{hyperref}       
\usepackage{url}            
\usepackage{booktabs}       
\usepackage{amsfonts}       
\usepackage{microtype}      
\usepackage{amsmath}
\usepackage{mathtools}
\usepackage{algorithm}
\usepackage{algorithmic}
\usepackage[dvipsnames]{xcolor}
\usepackage{tikz}
\usepackage{pgfplots}
\pgfplotsset{compat=1.14}
\usepgfplotslibrary{fillbetween}
\usepackage[algo2e]{algorithm2e}
\usepackage{graphicx}

\title{Direct Federated Neural Architecture Search}

\author{%
  Anubhav Garg\thanks{Correspondence to \texttt{anubhgar@cisco.com}}  \\
  Cisco Systems\\

  \AND
   Amit Kumar Saha \\
   Cisco Systems\\

  \And
   Debo Dutta \\
   Cisco Systems\\

}
\begin{document}

\maketitle

\begin{abstract}
Neural Architecture Search (NAS) is a collection of methods to craft the way neural networks are built. We apply this idea to Federated Learning (FL), wherein predefined neural network models are trained on the client/device data. This approach is not optimal as the model developers can't observe the local data, and hence, are unable to build highly accurate and efficient models. NAS is promising for FL which can search for global and personalized models automatically for the non-IID data. Most NAS methods are computationally expensive and require fine tuning after the search, making it a two-stage complex process with possible human intervention. Thus there is a need for end-to-end NAS which can run on the heterogeneous data and resource distribution typically seen in a FL scenario. In this paper, we present an effective approach for direct federated NAS which is hardware agnostic, computationally lightweight, and a one-stage method to search for ready-to-deploy neural network models. Our results show an order of magnitude reduction in resource consumption while edging out prior art in accuracy. This opens up a window of opportunity to create optimized and computationally efficient federated learning systems. 
\end{abstract}

\section{Introduction}

Federated Learning (FL) (\citet{McMahan2017}) is an approach to solving a machine learning problem where many clients generate their own data and collaborate under the orchestration of a central server to train the model, often to preserve privacy of the original data (e.g., in regulated verticals like healthcare). Mathematically, it can be represented by the following optimization problem: 

\begin{equation}
  \label{eq1}
\min_w \, \mathcal{F}(w), \; \; \text{where}  \; \; \mathcal{F}(w) \coloneqq \; \sum^{K}_{i=1} \; \nu_i \mathcal{F}_i(w)
\end{equation}

Here $K$ is the total number of clients, $\nu_i$ $\geq$ $0$ and $\sum_{i}$ $\nu_i=1$. $\mathcal{F}_i$ is the local objective function for client $i$ and is typically taken as $\mathcal{L}(x_i, y_i, w, \alpha)$, that is, the loss of the prediction on local data $(x_i, y_i)$ with model parameters $w$ and architecture $\alpha$. 
The term $\nu_i$ can be considered as the relative effect of each client, and is user defined with two typical values being $\nu_i$ $=$ $1/n$ or $\nu_i$ $=$ $n_i/n$ 
where $n_i$ is the number of samples of client $i$ and $n$ $=$ $\sum_{i}$ $n_i$.

A unique characteristic of FL is that each client's data is private to itself and not exchanged or transferred either with the server or other clients. Here, clients can be organizations like hospitals which have patients' data, such as CT scans
or edge devices like smart cameras or smartphones. 
We'll use the term clients in this article for simplicity, without loss of generality. Multiple such clients generate data locally, mostly in a non identical and independently distributed (IID) fashion.

To design an efficient and accurate model, model developers need to exhaustively examine the characteristics of the data, which is not possible in FL due to the invisibility of data from all the clients to the machine learning engineers. The current practice in FL is to apply a predefined neural network architecture for the given task. It involves many iterations of model selection and hyperparameter tuning requiring many rounds of training, which is extremely expensive and difficult to achieve on the limited computational resources and low communication bandwidth scenarios. This makes the objective function in Eq. \ref{eq1} hard to optimize. The objective in Eq. \ref{eq1} trains a global model for all the clients. If the clients' data is small and IID, the single global model learnt via federated learning should perform better than local models. A core challenge in FL is that the clients continuously generate data in a non-IID manner (\citet{cmu_fl_compre}), which makes a single predefined model hard to fit for all clients. Moreover, the predefined architecture design may have some modules which are redundant for a particular client's dataset and could lead to useless computation for that client.

To solve this problem, we make the use of Neural Architecture Search (NAS), which automatically searches for the optimal neural network architecture for the given task.
There are many search methods in NAS, majority of which can be classified as using reinforcement learning, evolutionary algorithms, and gradient descent. Compared to
other approaches, gradient based differentiable architecture search methods are order of magnitude cheaper with state-of-the-art performance. 
These methods employ a one-shot weight-shared model trained using gradient descent. Most of these methods involve a search stage, in which a child architecture is derived from the one-shot model and a retraining stage, in which a bigger network based on the child architecture is trained on the specified dataset. 
Thus, it is a two-stage process which involves analyses of the target data for building the final network. For example, DARTS (\citet{liu2018darts}) based methods search for a cell using a proxy dataset in the architecture search phase. For best performance, this cell is stacked many times according to the final dataset and retrained on it. Moreover, due to the bi-level optimization of DARTS, it is extremely computationally expensive and not fit for edge devices like smartphones (see section  \ref{results}, \ref{discussion} for further analysis). These methods are developed for centralized training and cannot be used directly in the FL setting. To address these issues, we propose an end-to-end NAS approach developed for the FL setting which can search for the optimal architecture directly on the given task, data, and hardware. It is a one-stage hardware adaptable method which searches for a ready-to-deploy neural network on any type of client hardware, ranging from hospitals having multi-GPU clusters to edge devices like smartphones or other IoT devices. 
The main contributions of our work can be summarized as follows:
\begin{enumerate}
    \item To the best of our knowledge, it is the first work to combine federated NAS for both cross-silo and cross-device environment, where the data is partitioned by examples by taking the advantage of \emph{Shufflenet} blocks (\citet{ShuffleNetAE}) in our search space, which is an extremely computation efficient architecture for mobile devices.
    \item We provide a efficient plug-and-play solution to search for a ready-to-deploy network in the federated setting. Our one-stage architecture search method
    eliminates the need for retraining of the derived network on the clients.
    \item As shown by extensive experiments, our approach takes significantly less communication rounds and computation resources compared to previous methods. For the same number of communication rounds, our approach achieves an average test accuracy improvement of upto $10\%$ compared to predefined models with $\mathrm{FedAvg}$ (\citet{McMahan2017}).
\end{enumerate}

\section{Our Approach}

In this section, we first formally define the problem, and then describe the construction
of the super network which includes any possible network in the search space. 
In Section \ref{ch}, we explain an efficient approach to search for a sub-network in this super network. Finally, we propose the direct federated NAS technique for target hardware which to optimize the objective in Eq. \ref{eq1}.
\subsection{Problem Definition}

As mentioned earlier, the definition of a typical federated learning problem is given by
Eq. (\ref{eq1}). In previous work, the architecture $\alpha$ of the model is fixed and the optimization is only on model parameters $w$, as can be seen from Eqn.~(\ref{eq1}). 
In contrast, we learn the architecture along with the model weights, 
$\alpha$ being a variable parameter. 
The local objective function $\mathcal{F}_i$ of each client therefore is (\citet{xie2018snas}):
\begin{equation}
  \label{eq2}
\mathbb{E}_{Z\sim p_{\alpha(Z)}} [\mathcal{L}_w(x_i, y_i, Z)]
\end{equation}

Here $Z$ is a matrix of all structural decisions, $\alpha$ is the architecture encoding, $p_\alpha(Z)$ is the distribution of architectures, and $\mathcal{L}$ is the loss function.
In other words, the local objective is to find the best $\alpha$ to optimize the expected performance on the given dataset $(x_i, y_i)$. The global objective is to maximize the predictive performance on the given task by collectively learning the model parameters along with the architecture $\alpha$.
\subsection{Direct Federated NAS}
\subsubsection{Preliminaries}
The objective function (Eqn.~(\ref{eq2})) can be optimized by searching on a \emph{supernet}, called the parent network. The parent network is a directed acyclic graph (DAG) \( \mathcal{G} =  ( \mathcal{V}, \mathcal{E} ) \). Nodes $v$ $\in$ \( \mathcal{V} \) represents the intermediate representations and edges $e$ $\in$ \( \mathcal{E} \) represent some transformation of them (e.g., pooling, convolution). There are multiple edges $e_{ij}$ between pair of nodes $v_i$ and node $v_j$ to comprise all possible architectures in the search space. Every edge has a weight $\alpha$, called the architecture parameter. These architecture parameters are trained by a NAS algorithm and for every pair of nodes, the architecture weight with the highest value is chosen to derive the final network, also called the child network.
\subsubsection{Searching for Child Network}
\label{ch}
Our architecture search method is based on DSNAS (\citet{Hu2020DSNASDN}). It is a task-specific end-to-end NAS method which searches for the child network in one stage. Algorithm~\ref{alg:1} details the specifics of searching for the final network.
\begin{algorithm}[H]
 \caption{Client Local Search}\
  \label{alg:1}
\SetAlgoLined
Require : parent network, operation parameters $w$, $\alpha_{th}$, total epochs $E$ \newline
 \For{\textnormal{epoch} $e = 1, 2, 3, ..., E $ \,}
 {
  1. Sample one-hot random variables $Z$ from $p_\alpha(Z)$ \newline
  2. Construct child network with $w$ according to $Z$, multiply a $\mathbf{1}$ after each feature map X \newline
  3. Get a batch from data and forward to get $\mathcal{L}$ \newline
  4. Backward $\mathcal{L}$ to both $w$ and $\mathbf{1}$, backward $\log p_\alpha(Z)$ to $\alpha$ \newline
  5. Update $w$ with $\frac{\partial \mathcal{L}}{\partial w}$, update $\alpha$ with $ \frac{\partial \log p_\alpha(z)}{\partial \alpha}$ $\frac{\partial \mathcal{L}}{\partial \mathbf{1}}  $ \newline
  6. Prune edges with weight $\alpha < \alpha_{th}$
 }
\end{algorithm}
The algorithm starts by sampling from $p_\alpha(Z)$, $Z$ is a matrix whose rows $Z_{i,j}$ are one-hot random variable vectors indicating masks multiplied to edges $e_{i, j}$ in the DAG. The probability distribution $p_\alpha(Z)$ is the softmax over $\alpha$. A child network is constructed based on $Z$ and the loss is calculated. The gradients are calculated using backpropagation and the weights as well as the architecture parameters are updated according to the formulae in Algorithm~\ref{alg:1}. The operation with the highest architecture parameter is selected between any pair of nodes.
\subsubsection{Federated NAS}
The objective in Eq. \ref{eq1} can be optimized by updating the global architecture and weight parameters, $\alpha$ and $w$ respectively by averaging the clients' values (\citet{McMahan2017}. Algorithm~\ref{alg:2} specifies such an orchestration by a central server.
\begin{algorithm}[H]
 \caption{Direct Federated NAS}\
  \label{alg:2}
\SetAlgoLined
 Require : Hardware configuration, the total number of rounds $T$ \\
 Build the parent network based on hardware configuration \\ Initialize $w$, $\alpha$ of parent network. Call it $w_0$ and $ \alpha_0$ \newline
 \For{round $t = 0, 1, 2, ..., T$ }{
  Select $K$ clients $S_t$ \\
 \For{ \textnormal{each client} $i \in S_t$ \bf {in parallel} }{
  Send $w_t$, $\alpha_t$ to client. Run Algorithm~\ref{alg:1} and get updated parameters, $w_{t+1}^i$ and $\alpha_{t+1}^i$
  }
  Update architecture $\alpha_{t+1}$ $\gets$ $\sum_{i}^{K} \frac{n_i}{n} \alpha_{t}^{i}$ \newline
    Update weight $w_{t+1}$ $\gets$ $\sum_{i}^{K}$ $\frac{n_i}{n} w_{t}^{i}$
}
\end{algorithm}
The above algorithm requires the device type as input, which can be GPU or CPU, and the parent network is constructed accordingly. The server initializes the weight and architecture parameters of the parent network. For every round, the server selects $K$ clients from the pool of available clients. In each round, the the local search process is run by every client in parallel and the architecture and weight parameters are trained on their respective dataset. After the clients have finished the local search, the architecture $\alpha$ and  weight $w$ parameters of the parent network is updated by the formulae in Algorithm~\ref{alg:2}. This process is repeated for all rounds or until the model is converged. \\
This method gives a ready-to-deploy network with no need of retraining a larger network constructed by stacking the DAG's learnt on a small network. Our method is hardware agnostic, which can search on the target hardware type directly by constructing the parent network accordingly. It learns the architecture directly on the federated datasets and target hardware without any proxy or manual intervention.

\section{Experiments and Results}
\label{results}
To evaluate our approach, we do several experiments for the cross-silo and cross-device setting. We take the target task as image classification in our experiments.  We use the FedML (\citet{fedml2020}) research library for our experiments.
\paragraph{Search space:} Our parent network is built using shuffle blocks (\citet{ShuffleNetAE}). For the cross-silo evaluation, there are 4 choices for each block in the parent network, with kernel sizes 3, 5, 7, and a xception block. For the cross-device experiments, there are 3 candidates for each choice block in the parent network. For further details about our search space, we refer to (\citet{Hu2020DSNASDN}). Our search space comprises of $4^{20}$ neural networks architectures for the cross-silo setting and $3^{12}$ for the cross-device setting.
 \paragraph{Dataset :} For cross-silo setting, we perform our experiments on the CIFAR-10 dataset for both IID and non-IID distribution. For the IID case, we divide the training dataset homogeneously between the clients in each round. We generate the non-IID dataset similar to (\citet{He2020FedNASFD}), for fair comparison, that is, by splitting the training images into $K$ clients in an unbalanced manner: sampling $p_c$ $\sim$ Dir($\alpha$) where Dir is the dirichlet distribution with $\alpha=0.5$ and allocating a $p_{c,k}$ proportion of the samples of class $c$ to local client $k$. Similar to (\citet{He2020FedNASFD}) we test different methods on the test data held by the central server. CIFAR-10 has 50,000 training and 10,000 test images,
 all coloured, of size 32x32, and split equally in 10 classes.
 
 For the cross-device setting, we use the CINIC-10 dataset, which has 90,000 training and test coloured images of size 32x32 split equally in 10 classes. The sampling strategy is same as that of the cross-silo setting.

\paragraph{Results:} Table \ref{tab:method_compar} compares the test accuracy and search cost of our method with FedNAS, which applies MiLeNAS (\citet{milenas}) architecture search for federated learning. We achieve remarkable computational efficiency improvement. As our method is one-stage, we record the total time taken to search and train. We would like to point out that though the parameters of the final network is more for our approach, the memory consumption during search is orders of magnitude lesser

\begin{figure}[htp]
\begin{minipage}[htp]{.45\textwidth}
\centering
\label{fig:validation_accuracy}
\begin{tikzpicture}
\centering
    \begin{axis}[
            xlabel={Communication Rounds},
            ylabel={Test Accuracy},
            grid style=dashed,
            legend entries={{crosses},{line/dots}},
            xmin=-5,
            legend style={legend pos=south east,font=\scriptsize},
            height=1.\textwidth,
            width=1.\textwidth,
        ]
        \addplot[
            color=orange,
            ]
        table {fedavg.dat};
        \addlegendentry{Mobilenet}
        \addplot[
            color=red,
            ]
        table {fedavg_resnet56.dat};
        \addlegendentry{Resnet56}
        \addplot[
            color=green
            ]
        table {fednas.dat};
        \addlegendentry{DFNAS}
    \end{axis}
\end{tikzpicture}
\caption{Average test accuracy vs. communication rounds for the non-IID CIFAR-10 data distribution on 8 clients for 150 rounds. DFNAS achieves an accuracy of $88.31\%$, MobileNetV1 $83.41\%$, and ResNet56 $77.78\%$ with FedAvg for same number of rounds.}
\end{minipage}
\hspace{1.1cm}
\begin{minipage}[htp]{.45\textwidth}
    \centering
    \begin{tikzpicture}
    \centering
        \begin{axis}[
            xlabel={Communication Rounds},
            ylabel={Test Accuracy},
            grid style=dashed,
            legend entries={{crosses},{line/dots}},
            xmin=-10,
            legend style={legend pos=south east,font=\scriptsize},
            height=1\textwidth,
            width=1\textwidth,
        ]
        \addplot[
            color=orange,
            ]
        table {shufflenet.dat};
        \addlegendentry{ShuffleNet}

        \addplot[
            color=green
            ]
        table {mobile.dat};
        \addlegendentry{DFNAS}
    \end{axis}
    \end{tikzpicture}
    \caption{Average test accuracy vs. communication rounds for the non-IID CINIC-10 data distribution on 16 clients for 250 rounds. DFNAS achieves an accuracy of $68.35\%$, while ShuffleNetV1 with FedAvg $67.70\%$  for same number of rounds.}
    \end{minipage}
\end{figure}
 making our approach fit for low memory hardware.
Table \ref{tab:clients_compar} compares the accuracy with different number of clients and time taken. It can be seen that the approach is robust to different clients and scale.

Figure 1 compares our approach with FedAvg algorithm on two pre-defined neural networks. DFNAS achieves more than $10\%$ higher test accuracy on Resnet56 (\citet{resnet}) and $5\%$ on MobileNet (\citet{MobileNets}) for same number of communication rounds. The lower accuracy of DFNAS for a few rounds is because it searches for the best architecture from the search space while for other curves it is fixed. 
Figure 2 compares DFNAS with \emph{ShuffleNetV1} using the FedAvg algorithm on the cross-device setting. The total number of clients are 16, both the methods have equal parameters ($< 1M$) and the data distribution is non-IID. For same number of communication rounds, we achieve better performance than the pre-defined architecture.
\begin{table}[htbp]
    \caption{DFNAS achieves state-of-the art performance on CIFAR-10 dataset. The results are for 8 clients having non-IID data distribution. Since our method is one-stage, the total time consisting of search and training is recorded.}
    \centering
    \begin{tabular}{c|c|c|c|c}
    \toprule
    Method & Test Accuracy (\%) & Params (M) & Total time (GPU Days) & Memory (MB) \\
    \midrule
    FedNAS       & 91.43 $\pm$ 0.13 & 0.33 & 1  & 10,793\\
    \hline
    DFNAS (ours) & 92.11 $\pm$ 0.1 &  2.1 & 0.18 &1,437\\
    \bottomrule
    \end{tabular}
    \label{tab:method_compar}
\end{table}
\begin{table}[htbp]
  \caption{Comparison of our approach with different clients on the CIFAR-10 dataset.  The data distribution is IID for more than one clients.}
  \label{search_cost_compare}
  \centering
  \begin{tabular}{c|c|c|c}
    \toprule
    Clients & Test Accuracy (\%) & Params (M) & Total time (GPU Days)\\
    \midrule
    2 & 93.53 $\pm$  0.10  & 2.1 & 0.83   \\
    \hline
    4     & 93.31 $\pm$  0.12 & 2.1 & 0.43     \\
    \hline
    8     & 92.79 $\pm$  0.15 & 2.1 & 0.18 \\
    \bottomrule
  \end{tabular}
  \label{tab:clients_compar}
\end{table}

\section{Related Work}
\paragraph{Federated Learning:} The approach to train shared model on decentralized data without transferring or exchanging it by aggregating locally computed updates was termed \emph{Federated Learning} (\citet{McMahan2017}). They introduced the Federated Averaging algorithm, in which the clients use SGD to train the shared model weights which are averaged by the central server. There are four core challenges in achieving the objective in Eq. \ref{eq1} --- statistical heterogeneity of data, hardware heterogeneity of clients, low communication bandwidth, and privacy (\citet{cmu_fl_compre}). Our method solves for first two of the above mentioned challenges directly and it solves the third challenge indirectly by eliminating the need for large number of communication rounds required due to model refinement. We would like to point out that standard privacy preserving approaches like differential privacy can be applied to our method to preserve
server and client privacy.

\paragraph{Neural Architecture Search:} \citet{nasnet} used a controller RNN and trained it with reinforcement learning to search for architectures. Since then, numerous NAS methods have been studied. Based on major search strategy, NAS methods can be classified into Reinforcement Learning (\citet{metaqnn, network-transformation, block-nas, nasnet-cell, enas}), Neuro-Evolution (\citet{real2017large, suganuma2018exploiting, hierarchical-evolution, real2019regularized, lamarckian}) and gradient-based (\citet{liu2018darts, proxylessnas, xie2018snas, xnas, pcdarts, progressive-darts}). Other methods include Random Search (\citet{random-search}), Bayesian Optimization (\citet{jin2019auto, bayesian-nas} and some custom methods (\citet{renas, envelopenet, manas, progressive-nas}).

\paragraph{Federated Neural Architecture Search (FNAS): } There are several works on FNAS which apply some NAS algorithm to the federated learning setting. In  (\citet{He2020FedNASFD}), clients use MiLeNAS (\citet{milenas}) to search for an architecture for the cross-silo FL setting. In DP-FNAS (\citet{Singh2020DPFNAS}), in each round clients compute weight and architecture parameter gradients on a subset of local data which are sent to server for averaging in two steps, thus requiring double communication rounds than our approach. DP-FNAS utilize DARTS with additionally applying random Gaussian noise to local gradients to preserve differential privacy. Though they achieve differential privacy by adding random noise to gradients, there is a tradeoff between validation accuracy and the level of privacy. Both of these approaches have large memory requirement than our approach. They are two-stage methods which require parameter retraining after the architecture search, thus requiring large number of communication rounds. 
DecNAS (\citet{Xu2020FNAS}) applies model compression and pruning techniques to a pre-trained model based on the NetAdapt (\citet{Yang2018NetAdaptPN}) framework for fitting on the given resource budget. Concretely, they prune the convolution filters with the smallest value based on the $\ell_{2}$ norm. DecNAS searched architecture has a performance limitation as that of the pre-trained model used with a key challenge to fit on the non-IID data in the FL setting. (\citet{Zhu2020RealtimeFE}) proposes a multi-objective double sampling evolutionary approach to FNAS which is computationally expensive.

\section{Discussions}
\label{discussion}

The computational complexity of our method is the same as training a single neural network since we sample only one path in the parent supernet DAG at every step instead of considering all the candidate paths, which requires the whole network to be stored in the memory. Specifically, if $P$, $Q$ and $R$ denote the forward time, backward time and memory consumption of a sampled subnetwork from the search space, then DARTS-like methods take $nP$, $nQ$ and $nR$, where $n$ is the number of operations while DFNAS takes the same as a single network, that is, $P$, $Q$ and $R$.

\section{Conclusions}
In this paper, we addressed the inefficiency of the current practice of applying predefined neural network architecture to federated learning systems. We presented a plug-and-play solution DFNAS, which is as simple as training one single neural network yet provides an improvement over current approaches in terms of accuracy, computation and communication bandwidth. 
Our work highlights the inefficiency of applying the DARTS based NAS methods directly on the federated learning setting. We believe that this general approach could be applied to a variety of federated learning problems. 

\bibliography{references}
\bibliographystyle{bib_style}
\end{document}